\documentclass[11pt,a4paper]{article}
\usepackage[hyperref]{emnlp-ijcnlp-2019}
\usepackage{times}
\usepackage{latexsym}
\usepackage{xcolor}
\usepackage{algorithm2e}
\usepackage{booktabs}
\usepackage{algorithmic}
 \usepackage{graphicx}
\usepackage{tikz}
\usepackage{pgfplots}
\usepackage{rotating}
\usepackage{enumitem}

\usepackage{graphicx}
\usepackage{bm}
\usepackage{amsmath,amsfonts,amssymb}
\usepackage{subcaption}
\usepackage{comment}

\pgfplotsset{compat=1.14}

\aclfinalcopy 

\newcommand{\cl}[1]{{\bf \color{blue} [CL: #1]}}

\newcommand{\lc}[1]{{\bf \color{blue} [LC: #1]}}
\newcommand{\claimNum}{43837}
\newcommand{\docNum}{257982}

\title{MultiFC: A Real-World Multi-Domain Dataset for\\ Evidence-Based Fact Checking of Claims}
  
\author{
Isabelle Augenstein \hspace{0.5cm} Christina Lioma \hspace{0.5cm} Dongsheng Wang \hspace{0.5cm} Lucas Chaves Lima \\ \textbf{Casper Hansen} \hspace{0.5cm} \textbf{Christian Hansen} \hspace{0.5cm} \textbf{Jakob Grue Simonsen} \\
	    Department of Computer Science\\
	    University of Copenhagen\\
	    {\tt \{augenstein,c.lioma,wang,lcl,c.hansen,chrh,simonsen\}@di.ku.dk}\\}

\date{}

\begin{document}
\maketitle

\begin{abstract}

We contribute the largest publicly available dataset of naturally occurring factual claims for the purpose of automatic claim verification. It is collected from 26 fact checking websites in English, paired with textual sources and rich metadata, and labelled for veracity by human expert journalists.
We present an in-depth analysis of the dataset, highlighting characteristics and challenges. 
Further, we present results for automatic veracity prediction, both with established baselines and with a novel method for joint ranking of evidence pages and predicting veracity that outperforms all baselines. Significant performance increases are achieved by encoding evidence, and by modelling metadata.
Our best-performing model achieves a Macro F1 of 49.2\%, showing that this is a challenging testbed for claim veracity prediction.

\end{abstract}


\section{Introduction}
\label{s:intro}

Misinformation and disinformation are two of the most pertinent and difficult challenges of the information age, exacerbated by the popularity of social media. In an effort to counter this, a significant amount of manual labour has been invested in fact checking claims, often collecting the results of these manual checks on fact checking portals or websites such as politifact.com or snopes.com. In a parallel development, researchers have recently started to view fact checking as a task that can be partially automated, using machine learning and NLP to automatically predict the \textit{veracity} of claims. 
However, existing efforts either use small datasets consisting of naturally occurring claims (e.g. \citet{mihalcea2009lie,zubiaga2016analysing}), or datasets consisting of artificially constructed claims such as FEVER \cite{thorne-EtAl:2018:N18-1}. While the latter offer valuable contributions to further automatic claim verification work, they cannot replace real-world datasets.


\begin{table}[]
\fontsize{10}{10}\selectfont
\begin{tabular}{p{2.1cm}p{4.7cm}}   
\toprule
Feature & Value \\ \midrule
ClaimID & farg-00004 \\
Claim & Mexico and Canada assemble cars with foreign parts and send them to the U.S. with no tax. \\
Label & distorts \\
Claim URL & \scriptsize{\url{https://www.factcheck.org/2018/10/factchecking-trump-on-trade/}} \\
Reason & None \\
Category & the-factcheck-wire \\
Speaker & Donald Trump \\
Checker & Eugene Kiely \\
Tags & North American Free Trade Agreement \\
Claim Entities & United\_States, Canada, Mexico \\
Article Title & FactChecking Trump on Trade\\
Publish Date & October 3, 2018 \\
Claim Date & Monday, October 1, 2018 \\ 
\bottomrule
\end{tabular}
\caption{\label{tb:examp_claim} An example of a claim instance. Entities are obtained via entity linking. Article and outlink texts, evidence search snippets and pages are not shown.}
\end{table}


\paragraph{Contributions.} We introduce the currently largest claim verification dataset of naturally occurring claims.\footnote{The dataset is found here: \url{https://copenlu.github.io/publication/2019_emnlp_augenstein/}} It consists of 34,918 claims, collected from 26 fact checking websites in English; evidence pages to verify the claims; the context in which they occurred; and rich metadata (see Table \ref{tb:examp_claim} for an example).
We perform a thorough analysis to identify characteristics of the dataset such as entities mentioned in claims. We demonstrate the utility of the dataset by training state of the art veracity prediction models, and find that evidence pages as well as metadata significantly contribute to model performance. Finally, we propose a novel model that jointly ranks evidence pages and performs veracity prediction. The best-performing model achieves a Macro F1 of 49.2\%, showing that this is a non-trivial dataset with remaining challenges for future work.



\section{Related Work}
\label{s:rw}

\begin{table*}[!htbp]
\fontsize{10}{10}\selectfont
    \centering
   \begin{tabular}{l c c c c c c c}
\toprule
\bf Dataset & \bf \# Claims & \bf Labels & \bf metadata & \bf Claim Sources\\
\midrule
\multicolumn{3}{l}{\bf I: Veracity prediction w/o evidence} & & & \\
\citet{P17-2067} & 12,836 & 6  & Yes & Politifact\\
\citet{C18-1287}& 980 & 2  & No & News Websites \\
\midrule
\multicolumn{3}{l}{\bf II: Veracity} & &  & \\
\citet{bachenko2008verification} & 275 & 2 & No & Criminal Reports \\
\citet{mihalcea2009lie}& 600 & 2 & No & Crowd Authors \\
\citet{mitra2015credbank}$\dagger$ & 1,049 & 5 & No & Twitter \\
\citet{ciampaglia2015computational}$\dagger$ & 10,000 & 2 &  No & Google, Wikipedia \\
\citet{PopatMSW16} & 5,013 & 2  & Yes & Wikipedia, Snopes \\
\citet{2018arXiv180901286S}$\dagger$ &  23,921 & 2 & Yes & Politifact, gossipcop.com \\
Datacommons Fact Check\footnote{https://datacommons.org/factcheck/download} & 10,564 & 2-6 &  Yes & Fact Checking Websites \\
 \midrule
\multicolumn{3}{l}{\bf III: Veracity (evidence encouraged, but not provided)} & & \\
\citet{barron2018overview} & 150 & 3 &  No & factcheck.org, Snopes \\
\midrule
 \multicolumn{3}{l}{\bf IV: Veracity + stance} & & \\
\citet{vlachos2014fact}& 106 & 5 & Yes & Politifact, Channel 4 News \\
\citet{zubiaga2016analysing}& 330 & 3  & Yes & Twitter \\
\citet{derczynski2017semeval}& 325 & 3  & Yes & Twitter \\
\citet{DBLP:conf/naacl/BalyMGMMN18} & 422 & 2  & No & ara.reuters.com, verify-sy.com\\
\citet{thorne-EtAl:2018:N18-1}$\dagger$& 185,445 & 3 & No & Wikipedia \\
\midrule \midrule
  \multicolumn{3}{l}{\bf V: Veracity + evidence relevancy} & & \\
MultiFC & 36,534 & 2-40 & Yes & Fact Checking Websites & \\
\bottomrule
\end{tabular}
    \caption{\label{tab:Datasets} Comparison of fact checking datasets. $\dagger$ indicates claims are not `naturally occuring': \citet{mitra2015credbank} use events as claims; \citet{ciampaglia2015computational} use DBPedia tiples as claims; \citet{2018arXiv180901286S} use tweets as claims; and \citet{thorne-EtAl:2018:N18-1} rewrite sentences in Wikipedia as claims.}
\end{table*}

\subsection{Datasets}

Over the past few years, a variety of mostly small datasets related to fact checking have been released. An overview over core datasets is given in Table \ref{tab:Datasets}. 
The datasets can be grouped into four categories (I--IV). Category I contains datasets aimed at testing how well the veracity\footnote{We use \emph{veracity}, \emph{claim credibility}, and \emph{fake news} prediction interchangeably here -- these terms are often conflated in the literature and meant to have the same meaning.} of a claim can be predicted using the claim alone, without context or evidence documents.
Category II contains datasets bundled with documents related to each claim --  either topically related to provide context, or serving as evidence. Those documents are, however, not annotated.
Category III is for predicting veracity; they encourage retrieving evidence documents as part of their task description, but do not distribute them. 
Finally, category IV comprises datasets annotated for both veracity and stance. Thus, every document is annotated with a label indicating whether the document supports or denies the claim, or is unrelated to it. Additional labels can then be added to the datasets to better predict veracity, for instance by jointly training stance and veracity prediction models. 

Methods not shown in the table, but related to fact checking, are stance detection for claims \cite{DBLP:conf/naacl/FerreiraV16,PomerleauRao,augenstein-etal-2016-stance,kochkina-etal-2017-turing,augenstein-etal-2016-usfd,journals/ipm/ZubiagaKLPLBCA18,journals/corr/RiedelASR17}, satire detection \cite{W16-0802}, clickbait detection \cite{KGNK2017}, conspiracy news detection \cite{DBLP:journals/corr/TacchiniBVMA17}, rumour cascade detection \cite{Vosoughi1146} and claim perspectives detection \cite{chenseeing}.

Claims are obtained from a variety of sources, including Wikipedia, Twitter, criminal reports and fact checking websites such as politifact.com and snopes.com. The same goes for documents -- these are often websites obtained through Web search queries, or Wikipedia documents, tweets or Facebook posts.
Most datasets contain a fairly small number of claims, and those that do not, often lack evidence documents. An exception is \citet{thorne-EtAl:2018:N18-1}, who create a Wikipedia-based fact checking dataset. While a good testbed for developing deep neural architectures, their dataset is artificially constructed and can thus not take metadata about claims into account.

{\bf Contributions:} We provide a dataset that, uniquely among extant datasets, contains a large number of \emph{naturally occurring} claims 
and rich additional meta-information. 

\subsection{Methods}

Fact checking methods partly depend on the type of dataset used. Methods only taking into account claims typically encode those with CNNs or RNNs \cite{P17-2067,C18-1287}, and potentially encode metadata \cite{P17-2067} in a similar way. Methods for small datasets often use hand-crafted features that are a mix of bag of word and other lexical features, e.g.\ LIWC, and then use those as input to a SVM or MLP \cite{mihalcea2009lie,C18-1287,DBLP:conf/naacl/BalyMGMMN18}. Some use additional Twitter-specific features \cite{enayet2017niletmrg}. 
More involved methods taking into account evidence documents, often trained on larger datasets, consist of evidence identification and ranking following a neural model that measures the compatibility between claim and evidence \cite{thorne-EtAl:2018:N18-1,DBLP:conf/aaai/MihaylovaNMBMKG18,yin2018twowingos}. 

{\bf Contributions:} The latter category above is the most related to our paper as we consider evidence documents. However, existing models are not trained jointly for evidence identification, or for stance and veracity prediction, but rather employ a pipeline approach. 
Here, we show that a joint approach that learns to weigh evidence pages by their importance for veracity prediction can improve downstream veracity prediction performance.




\section{Dataset Construction}
\label{s:crawl}



We crawled a total of 43,837 claims with their metadata (see details in Table \ref{tb:stats4all}). 
We present the data collection in terms of selecting sources, crawling claims and associated metadata (Section \ref{ss:sources}); retrieving evidence pages; 
and linking entities in the crawled claims (Section \ref{ss:ent}).


\subsection{Selection of sources}
\label{ss:sources}
We crawled all active fact checking websites in English listed by Duke Reporters' Lab\footnote{\url{https://reporterslab.org/fact-checking/}} and on the Fact Checking Wikipedia page.\footnote{\url{https://en.wikipedia.org/wiki/Fact_checking}} This resulted in 38 websites in total (shown in Table \ref{tb:stats4all}). Out of these, ten websites could not be crawled, as further detailed in Table \ref{tb:not_cralwed_list}.
In the later experimental descriptions, we refer to the part of the dataset crawled from a specific fact checking website as a \textit{domain}, and we refer to each website as \textit{source}.

From each source, we crawled the ID, claim, label, URL, reason for label, categories, person making the claim (speaker), person fact checking the claim (checker), tags, article title, publication date, claim date, as well as the full text that appears when the claim is clicked. Lastly, the above full text contains hyperlinks, so we further crawled the full text that appears when each of those hyperlinks are clicked (outlinks).

There were a number of crawling issues, e.g. security protection of websites with SSL/TLS protocols, time out, URLs that pointed to pdf files instead of HTML content, or unresolvable encoding. In all of these cases, the content could not be retrieved.
For some websites, no veracity labels were available, in which case, they were not selected as domains for training a veracity prediction model. Moreover, not all types of metadata (category, speaker, checker, tags, claim date, publish date) were available for all websites; and availability of articles and full texts differs as well.

We performed semi-automatic cleansing of the dataset as follows. First, we double-checked that the veracity labels would not appear in claims. For some domains, the first or last sentence of the claim would sometimes contain the veracity label, in which case we would discard either the full sentence or part of the sentence. Next, we checked the dataset for duplicate claims. We found 202 such instances, 69 of them with different labels. Upon manual inspection, this was mainly due to them appearing on different websites, with labels not differing much in practice (e.g. `Not true', vs. `Mostly False'). We made sure that all such duplicate claims would be in the training split of the dataset, so that the models would not have an unfair advantage. Finally, we performed some minor manual merging of label types for the same domain where it was clear that they were supposed to denote the same level of veracity (e.g. `distorts', `distorts the facts').


This resulted in a total of 36,534 claims with their metadata.
For the purposes of fact verification, we discarded instances with labels that occur fewer than 5 times, resulting in 34,918 claims. The number of instances, as well as labels per domain, are shown in Table \ref{tab:results_per-domain} and label names in Table \ref{tb:dataset_labels_full} in the appendix. The dataset is split into a training part (80\%) and a development and testing part (10\% each) in a label-stratified manner.
Note that the domains vary in the number of labels, ranging from 2 to 27. Labels include both  straight-forward ratings of veracity (`correct', `incorrect'), but also labels that would be more difficult to map onto a veracity scale (e.g. `grass roots movement!', `misattributed', `not the whole story'). We therefore do not postprocess label types across domains to map them onto the same scale, 
and rather treat them as is. In the methodology section (Section \ref{s:exp}), we show how a model can be trained on this dataset regardless by framing this multi-domain veracity prediction task as a multi-task learning (MTL) one.

\subsection{Retrieving Evidence Pages}\label{ss:evidence}
\label{ss:retrieval}

The text of each claim is submitted verbatim as a query to the Google Search API (without quotes). The 10 most highly ranked search results are retrieved, for each of which we save the title; Google search rank; URL; time stamp of last update; search snippet; as well as the full Web page. We acknowledge that search results change over time, which might have an effect on veracity prediction. However, studying such temporal effects is outside the scope of this paper. Similar to Web crawling claims, as described in Section \ref{ss:sources}, the corresponding Web pages can in some cases not be retrieved, in which case fewer than 10 evidence pages are available.
The resulting evidence pages are from a wide variety of URL domains, though with a predictable skew towards popular websites, such as Wikipedia or The Guardian (see Table \ref{tb:domain_percent} for detailed statistics).

\subsection{Entity Detection and Linking}
\label{ss:ent}

\begin{figure}[!t]
\centering
\scalebox{0.30}{
\includegraphics[]{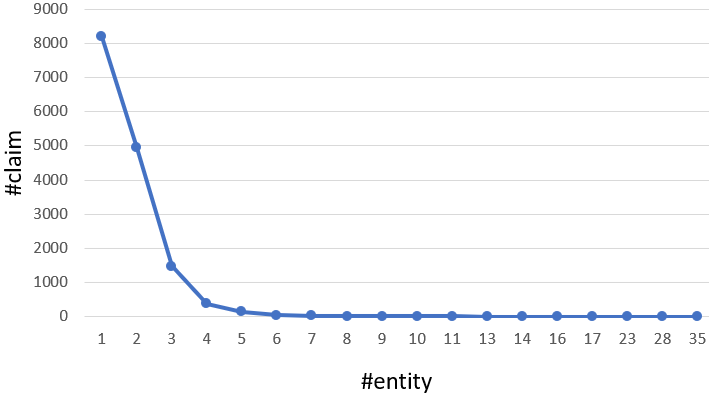}}
\caption{\label{fig:claim_distru4entity} Distribution of entities in claims. }
\end{figure}

To better understand what claims are about, we conduct entity linking 
for all claims. Specifically, mentions of people, places, organisations, and other named entities within a claim are recognised and linked to their respective Wikipedia pages, if available. Where there are different entities with the same name, they are disambiguated. For this, we apply the state-of-the-art neural entity linking model by \citet{kolitsas2018end}. 
This results
in a total of 25,763 entities detected and linked to Wikipedia, with a total of 15,351 claims involved, 
meaning that 42\% of all claims contain entities that can be linked to Wikipedia. Later on, we use entities as additional metadata (see Section \ref{ss:metadata}). 
The distribution of claim numbers according to the number of entities they contain is shown in Figure \ref{fig:claim_distru4entity}. We observe that the majority of claims have one to four entities, and the maximum number of 35 entities occurs in one claim only. Out of the 25,763 entities, 2,767 are unique entities. The top 30 most frequent entities are listed in Table \ref{tab:entiyFreq}. This clearly shows that most of the claims involve entities related to the United States, which is to be expected, as most of the fact checking websites are US-based.

\begin{table}[!t]
\fontsize{10}{10}\selectfont
\begin{tabular}{@{}lllll@{}}
\toprule
\bf Domain                          & \bf \%       &  &  &  \\ \midrule
https://en.wikipedia.org/       & 4.425 \\
https://www.snopes.com/         & 3.992 \\
https://www.washingtonpost.com/ & 3.025 \\
https://www.nytimes.com/        & 2.478 \\
https://www.theguardian.com/    & 1.807 \\
https://www.youtube.com/        & 1.712 \\
https://www.dailymail.co.uk/    & 1.558 \\
https://www.usatoday.com/       & 1.279 \\
https://www.politico.com/       & 1.241 \\
http://www.politifact.com/      & 1.231 \\
https://www.pinterest.com/      & 1.169 \\
https://www.factcheck.org/      & 1.09  \\
https://www.gossipcop.com/      & 1.073 \\
https://www.cnn.com/            & 1.065 \\
https://www.npr.org/            & 0.957 \\
https://www.forbes.com/         & 0.911 \\
https://www.vox.com/            & 0.89  \\
https://www.theatlantic.com/    & 0.88  \\
https://twitter.com/            & 0.767 \\
https://www.hoax-slayer.net/    & 0.655 \\
http://time.com/                & 0.554 \\
https://www.bbc.com/            & 0.551 \\
https://www.nbcnews.com/        & 0.515 \\
https://www.cnbc.com/           & 0.514 \\
https://www.cbsnews.com/        & 0.503 \\
https://www.facebook.com/       & 0.5   \\
https://www.newyorker.com/      & 0.495 \\
https://www.foxnews.com/        & 0.468 \\
https://people.com/             & 0.439 \\
http://www.cnn.com/             & 0.419                 \\ \bottomrule
\end{tabular}
\caption{\label{tb:domain_percent}The top 30 most frequently occurring URL domains.}
\end{table}

\begin{table}[!t]
\fontsize{10}{10}\selectfont
\begin{tabular}{lr}
\toprule
\textbf{Entity} & \textbf{Frequency} \\ 
\midrule
United\_States & 2810 \\
Barack\_Obama & 1598 \\
Republican\_Party\_(United\_States) & 783 \\
Texas & 665 \\
Democratic\_Party\_(United\_States) & 560 \\
Donald\_Trump & 556 \\
Wisconsin & 471 \\
United\_States\_Congress & 354 \\
Hillary\_Rodham\_Clinton & 306 \\
Bill\_Clinton & 292 \\
California & 285 \\
Russia & 275 \\
Ohio & 239 \\
China & 229 \\
George\_W.\_Bush & 208 \\
Medicare\_(United\_States) & 206 \\
Australia & 186 \\
Iran & 183 \\
Brad\_Pitt & 180 \\
Islam & 178 \\
Iraq & 176 \\
Canada & 174 \\
White\_House & 166 \\
New\_York\_City & 164 \\
Washington,\_D.C. & 164 \\
Jennifer\_Aniston & 163 \\
Mexico & 158 \\
Ted\_Cruz & 152 \\
Federal\_Bureau\_of\_Investigation & 146 \\
Syria & 130 \\ 
\bottomrule
\end{tabular}
\caption{\label{tab:entiyFreq} Top 30 most frequent entities listed by their Wikipedia URL with prefix omitted}
\end{table}

\section{Claim Veracity Prediction}
\label{s:exp}

We train several models to predict the veracity of claims. Those fall into two categories: those that only consider the claims themselves, and those that encode evidence pages as well. In addition, claim metadata (speaker, checker, linked entities) is optionally encoded for both categories of models, and ablation studies with and without that metadata are shown.
We first describe the base model used in Section \ref{ss:mtl}, followed by introducing our novel evidence ranking and veracity prediction model in Section \ref{ss:ranking}, and lastly the metadata encoding model in Section \ref{ss:metadata}.

\subsection{Multi-Domain Claim Veracity Prediction with Disparate Label Spaces}\label{ss:mtl}
Since not all fact checking websites use the same claim labels (see Table \ref{tab:results_per-domain}, and Table \ref{tb:dataset_labels_full} in the appendix), training a claim veracity prediction model is not entirely straight-forward. One option would be to manually map those labels onto one another. However, since the sheer number of labels is rather large (165), and it is not always clear from the guidelines on fact checking websites how they can be mapped onto one another, we opt to learn how these labels relate to one another as part of the veracity prediction model. To do so, we employ the multi-task learning (MTL) approach inspired by collaborative filtering presented in \citet{conf/naacl/AugensteinRS18} (\textit{MTL with LEL}--multitask learning with label embedding layer) that excels on pairwise sequence classification tasks with disparate label spaces. 
More concretely, each domain is modelled as its own task in a MTL architecture, and labels are projected into a fixed-length label embedding space. Predictions are then made by taking the dot product between the claim-evidence embeddings and the label embeddings. By doing so, the model implicitly learns how semantically close the labels are to one another, and can benefit from this knowledge when making predictions for individual tasks, which on their own might only have a small number of instances. When making predictions for individual domains/tasks, both at training and at test time, as well as when calculating the loss, a mask is applied such that the valid and invalid labels for that task are restricted to the set of known task labels.

Note that the setting here slightly differs from \citet{conf/naacl/AugensteinRS18}. There, tasks are less strongly related to one another; for example, they consider stance detection, aspect-based sentiment analysis and natural language inference. Here, we have different domains, as opposed to conceptually different tasks, but use their framework, as we have the same underlying problem of disparate label spaces.
A more formal problem definition follows next, as our evidence ranking and veracity prediction model in Section \ref{ss:ranking} then builds on it.

\subsubsection{Problem Definition}

We 
frame our problem as a multi-task learning one, where access to labelled datasets for $T$ tasks $\mathcal{T}_1, \ldots, \mathcal{T}_T$ is given at training time with a target task $\mathcal{T}_T$ that is of particular interest. The training dataset for task $\mathcal{T}_i$ consists of $N$ examples $X_{\mathcal{T}_i} = \{x_1^{\mathcal{T}_i}, \ldots, x_{N}^{\mathcal{T}_i}\}$ and their labels $Y_{\mathcal{T}_i} = \{\mathbf{y}_1^{\mathcal{T}_i}, \ldots, \mathbf{y}_{N}^{\mathcal{T}_i}\}$.
The base model is a classic deep neural network MTL model \cite{Caruana:93} that shares its parameters across tasks and has task-specific softmax output layers that output a probability distribution $\mathbf{p}^{\mathcal{T}_i}$ for task $\mathcal{T}_i$:

\begin{equation}
\mathbf{p}^{\mathcal{T}_i} = \mathrm{softmax}(\mathbf{W}^{\mathcal{T}_i}\mathbf{h} + \mathbf{b}^{\mathcal{T}_i})
\end{equation}

\noindent where $\mathrm{softmax}(\mathbf{x}) = e^\mathbf{x} / \sum^{ \|\mathbf{x}\| }_{i=1} e^{\mathbf{x}_i}$, $\mathbf{W}^{\mathcal{T}_i} \in \mathbb{R}^{L_i \times h}$, $\mathbf{b}^{\mathcal{T}_i} \in \mathbb{R}^{L_i}$ is the weight matrix and bias term of the output layer of task $\mathcal{T}_i$ respectively, $\mathbf{h} \in \mathbb{R}^h$ is the jointly learned hidden representation, $L_i$ is the number of labels for task $\mathcal{T}_i$, and $h$ is the dimensionality of $\mathbf{h}$.
The MTL model is trained to minimise the sum of individual task losses $\mathcal{L}_1 + \ldots + \mathcal{L}_T$ using a negative log-likelihood objective.






\paragraph{Label Embedding Layer.}

To learn the relationships between labels, a Label Embedding Layer (LEL) embeds labels of all tasks in a joint Euclidian space. Instead of training separate softmax output layers as above, a label compatibility function $c(\cdot, \cdot)$ measures how similar a label with embedding $\mathbf{l}$ is to the hidden representation $\mathbf{h}$:

\begin{equation}\label{eq:labelemb1}
c(\mathbf{l},\mathbf{h}) = \mathbf{l} \cdot \mathbf{h}
\end{equation}

\noindent where $\cdot$ is the dot product. Padding is applied such that $l$ and $h$ have the same dimensionality. 
Matrix multiplication and softmax are used for making predictions:
\begin{equation}
\mathbf{p} = \mathrm{softmax}(\mathbf{L} \mathbf{h})
\end{equation}

\noindent where $\mathbf{L} \in \mathbb{R}^{(\sum_i L_i) \times l}$ is the label embedding matrix for all tasks and $l$ is the dimensionality of the label embeddings. 
We apply a task-specific mask to $\mathbf{L}$ in order to obtain a task-specific probability distribution $\mathbf{p}^{\mathcal{T}_i}$. The LEL is shared across all tasks, which allows the model to learn the relationships between labels in the joint embedding space. 




\subsection{Joint Evidence Ranking and Claim Veracity Prediction}\label{ss:ranking}

\begin{figure}[!t]
      \centering
         \includegraphics[height=4.4in]{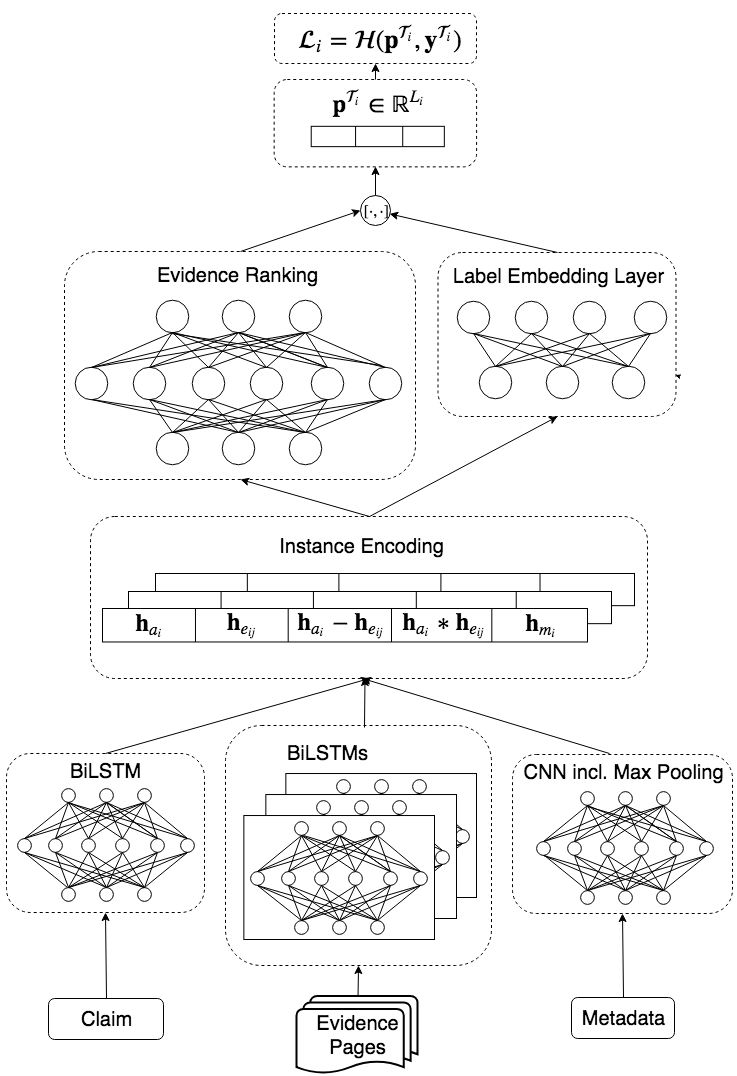}
    \caption{The Joint Veracity Prediction and Evidence Ranking model, shown for one task.}
\label{fig:training-procedures}
\end{figure}

So far, we have ignored the issue of how to obtain claim representation, as the base model described in the previous section is agnostic to how instances are encoded.
A very simple approach, which we report as a baseline, is to encode claim texts only. Such a model ignores evidence for and against a claim, and ends up guessing the veracity based on surface patterns observed in the claim texts.

We next introduce two variants of evidence-based veracity prediction models that encode 10 pieces of evidence in addition to the claim. Here, we opt to encode search snippets as opposed to whole retrieved pages. While the latter would also be possible, it comes with a number of additional challenges, such as encoding large documents, parsing tables or PDF files, and encoding images or videos on these pages, which we leave to future work. Search snippets also have the benefit that they already contain summaries of the part of the page content that is most related to the claim.

\subsubsection{Problem Definition}

Our problem is to obtain encodings for $N$ examples $X_{\mathcal{T}_i} = \{x_1^{\mathcal{T}_i}, \ldots, x_{N}^{\mathcal{T}_i}\}$. For simplicity, we will henceforth drop the task superscript and refer to instances as $X = \{x_1, \ldots, x_N\}$, as instance encodings are learned in a task-agnostic fashion.
Each example further consists of a claim $a_i$ and $k=10$ evidence pages $E_k = \{e_{1_{0}}, \ldots, e_{N_{10}}\}$.

Each claim and evidence page is encoded with a BiLSTM to obtain a sentence embedding, which is the concatenation of the last state of the forward and backward reading of the sentence, i.e. $\mathbf{h} = BiLSTM(\cdot)$, where $\mathbf{h}$ is the sentence embedding.

Next, we want to combine claims and evidence sentence embeddings into joint instance representations.
In the simplest case, referred to as model variant \textit{crawled\_avg}, we mean average the BiLSTM sentence embeddings of all evidence pages (signified by the overline) and concatenate those with the claim embeddings, i.e. 

\begin{equation}
\mathbf{s}_{{g}_i} = [ \mathbf{h}_{a_i};\overline{\mathbf{h}_{E_i}} ]
\end{equation}

\noindent where $s_{{g}_i}$ is the resulting encoding for training example $i$ and $[\cdot ; \cdot]$ denotes vector concatenation.
However, this has the disadvantage that all evidence pages are considered equal.

\paragraph{Evidence Ranking}
The here proposed alternative instance encoding model, \textit{crawled\_ranked}, which achieves the highest overall performance as discussed in Section \ref{s:results}, learns the compatibility between an instance's claim and each evidence page. It ranks evidence pages by their utility for the veracity prediction task, and then uses the resulting ranking to obtain a weighted combination of all claim-evidence pairs. No direct labels are available to learn the ranking of individual documents, only for the veracity of the associated claim, so the model has to learn evidence ranks implicitly.

To combine claim and evidence representations, we use the matching model proposed for the task of natural language inference by \citet{conf/acl/MouMLX0YJ16} and adapt it to combine an instance's claim representation with each evidence representation, i.e.

\begin{equation}
s_{{r}_{i_j}} = [ \mathbf{h}_{a_i};\mathbf{h}_{e_{i_j}};\mathbf{h}_{a_i} - \mathbf{h}_{e_{i_j}};\mathbf{h}_{a_i} \cdot \mathbf{h}_{e_{i_j}} ]
\end{equation}

\noindent where $s_{{r}_{i_j}}$ is the resulting encoding for training example $i$ and evidence page $j$ , $[\cdot ; \cdot]$ denotes vector concatenation, and $\cdot$ denotes the dot product.

All joint claim-evidence representations $\mathbf{s}_{{r}_{i_0}} , \ldots, \mathbf{s}_{{r}_{i_{10}}}$ are then projected into the binary space via a fully connected layer $\mathrm{FC}$, followed by a non-linear activation function $f$, to obtain a soft ranking of claim-evidence pairs, in practice a 10-dimensional vector,

\begin{equation}\label{eq:labelemb2}
\mathbf{o}_i = [ f(\mathrm{FC}( s_{{r}_{i_0}} )) ; \ldots ; f(\mathrm{FC}( s_{{r}_{i_{10}}} ))] 
\end{equation}

\noindent where $[\cdot ; \cdot]$ denotes concatenation.

Scores for all labels are obtained as per (\ref{eq:labelemb2}) above, with the same input instance embeddings as for the evidence ranker, i.e. $s_{{r}_{i_j}}$.
Final predictions for all claim-evidence pairs are then obtained by taking the dot product between the label scores and binary evidence ranking scores, i.e.

\begin{equation}\label{eq:labelemb}
\mathbf{p}_i = \mathrm{softmax}( c(\mathbf{l},\mathbf{s_{{r}_{i}}}) \cdot \mathbf{o}_i ) 
\end{equation}

\noindent Note that the novelty here is that, unlike for the model described in \citet{conf/acl/MouMLX0YJ16}, we have no direct labels for learning weights for this matching model. Rather, our model has to implicitly learn these weights for each claim-evidence pair in an end-to-end fashion given the veracity labels.

\begin{table}[!t]
\fontsize{10}{10}\selectfont
\begin{tabular}{@{}lcc@{}}
\toprule
\textbf{Model} & \textbf{Micro F1} & \textbf{Macro F1} \\ \midrule
claim-only & 0.469 & 0.253 \\
claim-only\_embavg & 0.384 & 0.302 \\
crawled-docavg & 0.438 & 0.248 \\
crawled\_ranked & 0.613 & 0.441 \\
\midrule
claim-only + meta & 0.494 & 0.324 \\
claim-only\_embavg + meta & 0.418 & 0.333 \\
crawled-docavg + meta & 0.483 & 0.286 \\
crawled\_ranked + meta & \bf 0.625 & \bf 0.492 \\
\bottomrule
\end{tabular}
\caption{\label{tab:results_main} Results with different model variants on the test set, `meta' means all metadata is used.}\end{table}

\subsection{Metadata}\label{ss:metadata}

We experiment with how useful claim metadata is, and encode the following as one-hot vectors: speaker, category, tags and linked entities. We do not encode `Reason' as it gives away the label, and do not include `Checker' as there are too many unique checkers for this information to be relevant. The claim publication date is potentially relevant, but it does not make sense to merely model this as a one-hot feature, so we leave incorporating temporal information to future work.
Since all metadata consists of individual words and phrases, a sequence encoder is not necessary, and we opt for a CNN followed by a max pooling operation as used in \citet{P17-2067} to encode metadata for fact checking.
The max-pooled metadata representations, denoted $h_m$, are then concatenated with the instance representations, e.g. for the most elaborate model, \textit{crawled\_ranked}, these would be concatenated with $s_{{cr}_{i_j}}$.

\section{Experiments}\label{s:results}

\subsection{Experimental Setup}

The base sentence embedding model is a BiLSTM over all words in the respective sequences with randomly initialised word embeddings, following \citet{conf/naacl/AugensteinRS18}. 
We opt for this strong baseline sentence encoding model, as opposed to engineering sentence embeddings that work particularly well for this dataset, to showcase the dataset. We would expect pre-trained contextual encoding models, e.g. ELMO \cite{conf/naacl/PetersNIGCLZ18}, ULMFit \cite{conf/acl/RuderH18}, BERT \cite{journals/corr/abs-1810-04805}, to offer complementary performance gains, as has been shown for a few recent papers \cite{conf/emnlp/WangSMHLB18,conf/acl/RajpurkarJL18}.

For claim veracity prediction without evidence documents with the MTL with LEL model, we use the following sentence encoding variants: \textit{claim-only}, which uses a BiLSTM-based sentence embedding as input, and \textit{claim-only\_embavg}, which uses a sentence embedding based on mean averaged word embeddings as input.

We train one multi-task model per task (i.e., one model per domain).
We perform a grid search over the following hyperparameters, tuned on the respective dev set, and evaluate on the correspoding test set (final settings are underlined): 
word embedding size [64, \underline{128}, 256], BiLSTM hidden layer size [64, \underline{128}, 256], number of BiLSTM hidden layers [1, \underline{2}, 3], BiLSTM dropout on input and output layers [0.0, \underline{0.1}, 0.2, 0.5], word-by-word-attention for BiLSTM with window size 10 \cite{bahdanau2014neural} [True, \underline{False}], skip-connections for the BiLSTM [\underline{True}, False], batch size [\underline{32}, 64, 128], label embedding size [\underline{16}, 32, 64]. We use ReLU as an activation function for both the BiLSTM and the CNN. 
For the CNN, the following hyperparameters are used: number filters [\underline{32}], kernel size [\underline{32}]. 
We train using cross-entropy loss and the RMSProp optimiser with initial learning rate of $0.001$ and perform early stopping on the dev set with a patience of $3$.

\subsection{Results}

\begin{table}[!t]
\fontsize{10}{10}\selectfont
\begin{tabular}{@{}lrccc@{}}
\toprule
\textbf{Domain} & \bf \# Insts &\bf  \# Labs & \textbf{Micro F1} & \textbf{Macro F1} \\ \midrule
ranz & 21 & 2 & 1.000 & 1.000 \\
bove & 295 & 2 & 1.000 & 1.000 \\
abbc & 436 & 3 & 0.463 & 0.453 \\
huca & 34 & 3 & 1.000 & 1.000 \\
mpws & 47 & 3 & 0.667 & 0.583 \\ 
peck & 65 & 3 & 0.667 & 0.472 \\
faan & 111 & 3 & 0.682 & 0.679 \\
clck & 38 & 3 & 0.833 & 0.619 \\ 
fani & 20 & 3 & 1.000 & 1.000 \\ 
chct & 355 & 4 & 0.550 & 0.513 \\
obry & 59 & 4 & 0.417 & 0.268 \\
vees & 504 & 4 & 0.721 & 0.425 \\
faly & 111 & 5 & 0.278 & 0.5 \\ 
goop & 2943 & 6 & 0.822 & 0.387 \\ 
pose & 1361 & 6 & 0.438 & 0.328 \\
thet & 79 & 6 & 0.55 & 0.37 \\
thal & 163 & 7 & 1.000 & 1.000 \\ 
afck & 433 & 7 & 0.357 & 0.259 \\
hoer & 1310 & 7 & 0.694 & 0.549 \\
para & 222 & 7 & 0.375 & 0.311 \\
wast & 201 & 7 & 0.344 & 0.214 \\
vogo & 654 & 8 & 0.594 & 0.297 \\ 
pomt & 15390 & 9 & 0.321 & 0.276 \\ 
snes & 6455 & 12 & 0.551 & 0.097 \\
farg & 485 & 11 & 0.500 & 0.140 \\ 
tron & 3423 & 27 & 0.429 & 0.046 \\ 
\midrule
avg &  & 7.17 & 0.625 & 0.492 \\ 
\bottomrule
\end{tabular}
\caption{\label{tab:results_per-domain} Total number of instances and unique labels per domain, as well as per-domain results with model \textit{crawled\_ranked + meta}, sorted by label size}
\end{table}

\begin{table}[!t]
\fontsize{10}{10}\selectfont
\begin{tabular}{@{}lcc@{}}
\toprule
\bf Metadata & \bf Micro F1 & \bf Macro F1 \\
\midrule
None & \bf 0.627 & 0.441 \\
\midrule
Speaker & 0.602 & 0.435 \\
  + Tags & 0.608 & 0.460 \\
\midrule
Tags & 0.585 & 0.461 \\
\midrule
Entity & 0.569 & 0.427 \\
  + Speaker & 0.607 & 0.477 \\
  + Tags & 0.625 & \bf 0.492 \\
\bottomrule
\end{tabular}
\caption{\label{tab:results_meta-ablation} Ablation results with base model \textit{crawled\_ranked} for different types of metadata}
\end{table}

\begin{table}[!t]
\fontsize{10}{10}\selectfont
\begin{tabular}{@{}lcc@{}}
\toprule
\bf Model & \bf Micro F1 & \bf Macro F1 \\
\midrule
STL & 0.527 & 0.388 \\
MTL & 0.556 & 0.448 \\
MTL + LEL & \bf 0.625 & \bf 0.492 \\
\bottomrule
\end{tabular}
\caption{\label{tab:results_training-ablation} Ablation results with \textit{crawled\_ranked + meta} encoding for STL vs. MTL vs. MTL + LEL training}
\end{table}

For each domain, we compute the Micro as well as Macro F1, then mean average results over all domains. Core results with all vs. no metadata are shown in Table \ref{tab:results_main}.
We first experiment with different base model variants and find that label embeddings improve results, and that the best proposed models utilising multiple domains outperform single-task models (see Table \ref{tab:results_training-ablation}). This corroborates the findings of \citet{conf/naacl/AugensteinRS18}.
Per-domain results with the best model are shown in Table \ref{tab:results_per-domain}. Domain names are from hereon after abbreviated for brevity, see Table \ref{tb:stats4all} in the appendix for correspondences to full website names. Unsurprisingly, it is hard to achieve a high Macro F1 for domains with many labels, e.g. tron and snes. Further, some domains, surprisingly mostly with small numbers of instances, seem to be very easy -- a perfect Micro and Macro F1 score of 1.0 is achieved on ranz, bove, buca, fani and thal. 
We find that for those domains, the verdict is often already revealed as part of the claim using explicit wording.

\paragraph{Claim-Only vs. Evidence-Based Veracity Prediction.}
Our evidence-based claim veracity prediction models outperform claim-only veracity prediction models by a large margin. Unsurprisingly, \textit{claim-only\_embavg} is outperformed by \textit{claim-only}. Further, \textit{crawled\_ranked} is our best-performing model in terms of Micro F1 and Macro F1, meaning that our model captures that not every piece of evidence is equally important, and can utilise this for veracity prediction.

\paragraph{Metadata.}

We perform an ablation analysis of how metadata impacts results, shown in Table \ref{tab:results_meta-ablation}. 
Out of the different types of metadata, topic tags on their own contribute the most. This is likely because they offer highly complementary information to the claim text of evidence pages. Only using all metadata together achieves a higher Macro F1 at similar Micro F1 than using no metadata at all. 
To further investigate this, we split the test set into those instances for which no metadata is available vs. those for which metadata is available. We find that encoding metadata within the model hurts performance for domains where no metadata is available, but improves performance where it is. In practice, an ensemble of both types of models would be sensible, as well as exploring more involved methods of encoding metadata.  



\begin{figure}[!t]
\centering
\scalebox{0.49}{
\includegraphics[]{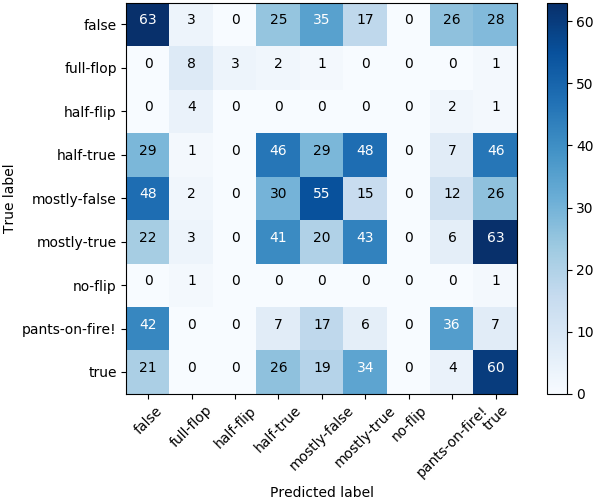}}
\caption{\label{fig:confusion_matrix} Confusion matrix of predicted labels with best-performing model, \textit{crawled\_ranked + meta}, on the `pomt' domain}
\end{figure}

\section{Analysis and Discussion}

An analysis of labels frequently confused with one another, for the largest domain `pomt' and best-performing model \textit{crawled\_ranked + meta} is shown in Figure \ref{fig:confusion_matrix}. The diagonal represents when gold and predicted labels match, and the numbers signify the number of test instances. One can observe that the model struggles more to detect claims with labels `true' than those with label `false'. Generally, many confusions occur over close labels, e.g. `half-true' vs. `mostly true'. 

We further analyse what properties instances that are predicted correctly vs. incorrectly have, using the model \textit{crawled\_ranked meta}.
We find that, unsurprisingly, longer claims are harder to classify correctly, and that claims with a high direct token overlap with evidence pages lead to a high evidence ranking.
When it comes to frequently occurring tags and entities, 
very general tags such as `government-and-politics' or `tax' that do not give away much, frequently co-occur with incorrect predictions, whereas more specific tags such as `brisbane-4000' or `hong-kong' tend to co-occur with correct predictions. 
Similar trends are observed for bigrams.
This means that the model has an easy time succeeding for instances where the claims are short, where specific topics tend to co-occur with certain veracities, and where evidence documents are highly informative. Instances with longer, more complex claims where evidence is ambiguous remain challenging.

\section{Conclusions}

We present a new, real-world fact checking dataset, currently the largest of its kind. It consists of 34,918 claims collected from 26 fact checking websites, rich metadata and 10 retrieved evidence pages per claim.
We find that encoding the metadata as well evidence pages helps, and introduce a new joint model for ranking evidence pages and predicting veracity. 

\section*{Acknowledgments}
This research is partially supported by QUARTZ (721321, EU H2020 MSCA-ITN) and DABAI (5153-00004A, Innovation Fund Denmark).

\bibliography{acl2019}
\bibliographystyle{acl_natbib}

\appendix

\setlength{\tabcolsep}{0.2em}
\begin{table*}[]
\fontsize{10}{12}\selectfont
\begin{tabular}{@{}lll@{}}
\toprule
 & \bf Websites (Sources)                                                           & \bf Reason                                                                                              \\ \midrule
 & Mediabiasfactcheck                                                           & Website that checks other news websites                                                             \\
 & CBC                                                                          & No pattern to crawl                                                                                 \\
 & apnews.com/APFactCheck                                                       & No categorical label and no structured claim                                                        \\
 & weeklystandard.com/tag/fact-check                                            & Mostly no label, and they are placed anywhere \\
 & ballotpedia.org                                                              & No categorical label and no structured claim                                                        \\
 & channel3000.com/news/politics/reality-check                                  & No categorical label, lack of structure, and no clear claim                                         \\
 & npr.org/sections/politics-fact-check & No label and no clear claim (only some titles are claims) \\
 & dailycaller.com/buzz/check-your-fact                                         & Is a subset of checkyourfact which has already been crawled                                         \\
 & sacbee.com\footnote{sacbee.com/news/politics-government/election/california-elections/poligraph/} & Contains very few labelled articles, and without clear claims                      \\
 & TheGuardian & Only a few websites have a pattern for labels. \\ \bottomrule
\end{tabular}
\caption{\label{tb:not_cralwed_list} The list of websites that we did not crawl and reasons for not crawling them.}
\end{table*}

\begin{table*}[]
\fontsize{10}{10}\selectfont
\begin{tabular}{p{1.1cm}p{1.0cm}p{1.3cm}p{11cm}}   
\toprule
\bf Domain & \bf \# Insts & \bf \# Labels & \bf Labels \\
\midrule
abbc & 436 & 3 & in-between, in-the-red, in-the-green \\
afck & 433 & 7 & correct, incorrect, mostly-correct, unproven, misleading, understated, exaggerated \\
bove & 295 & 2 & none, rating: false \\
chct & 355 & 4 & verdict: true, verdict: false, verdict: unsubstantiated, none \\
clck & 38 & 3 & incorrect, unsupported, misleading \\
faan & 111 & 3 & factscan score: false, factscan score: true, factscan score: misleading \\
faly & 71 & 5 & true, none, partly true, unverified, false \\
fani & 20 & 3 & conclusion: accurate, conclusion: false, conclusion: unclear \\
farg & 485 & 11 & false, none, distorts the facts, misleading, spins the facts, no evidence, not the whole story, unsupported, cherry picks, exaggerates, out of context \\
goop & 2943 & 6 & 0, 1, 2, 3, 4, 10 \\
hoer & 1310 & 7 & facebook scams, true messages, bogus warning, statirical reports, fake news, unsubstantiated messages, misleading recommendations \\
huca & 34 & 3 & a lot of baloney, a little baloney, some baloney \\
mpws & 47 & 3 & accurate, false, misleading \\
obry & 59 & 4 & mostly\_true, verified, unobservable, mostly\_false \\
para & 222 & 7 & mostly false, mostly true, half-true, false, true, pants on fire!, half flip \\
peck & 65 & 3 & false, true, partially true \\
pomt & 15390 & 9 & half-true, false, mostly true, mostly false, true, pants on fire!, full flop, half flip, no flip \\
pose & 1361 & 6 & promise kept, promise broken, compromise, in the works, not yet rated, stalled \\
ranz & 21 & 2 & fact, fiction \\
snes & 6455 & 12 & false, true, mixture, unproven, mostly false, mostly true, miscaptioned, legend, outdated, misattributed, scam, correct attribution \\
thet & 79 & 6 & none, mostly false, mostly true, half true, false, true \\
thal & 74 & 2 & none, we rate this claim false \\
tron & 3423 & 27 & fiction!, truth!, unproven!, truth! \& fiction!, mostly fiction!, none, disputed!, truth! \& misleading!, authorship confirmed!, mostly truth!, incorrect attribution!, scam!, investigation pending!, confirmed authorship!, commentary!, previously truth! now resolved!, outdated!, truth! \& outdated!, virus!, fiction! \& satire!, truth! \& unproven!, misleading!, grass roots movement!, opinion!, correct attribution!, truth! \& disputed!, inaccurate attribution! \\
vees & 504 & 4 & none, fake, misleading, false \\
vogo & 653 & 8 & none, determination: false, determination: true, determination: mostly true, determination: misleading, determination: barely true, determination: huckster propaganda, determination: false, determination: a stretch \\
wast & 201 & 7 & 4 pinnochios, 3 pinnochios, 2 pinnochios, false, not the whole story, needs context, none \\
\bottomrule
\end{tabular}
\caption{\label{tb:dataset_labels_full} Number of instances, and labels per domain sorted by number of occurrences}
\end{table*}

\begin{sidewaystable*}
\fontsize{9}{11}\selectfont
\begin{tabular}{llllllllllllll}
\toprule
\bf Website & \bf Domain & \bf Claims & \bf Labels & \bf Category & \bf Speaker & \bf Checker & \bf Tags & \bf Article & \bf Claim date & \bf Publish date & \bf Full text & \bf Outlinks \\ 
\midrule
abc & abbc & 436 & 436 & 436 & - & - & 436 & 436 & - & 436 & 436 & 7676 \\
africacheck & afck & 436 & 436 & - & - & - & - & 436 & - & 436 & 436 & 2325 \\
altnews & - & 496 & - & - & - & 496 & - & 496 & - & 496 & 496 & 6389 \\
boomlive & - & 302 & 302 & - & - & - & - & 302 & - & 302 & 302 & 6054 \\
checkyourfact & chht & 358 & 358 & - & - & 358 & - & - & - & 358 & 358 & 5271 \\
climatefeedback & clck & 45 & 45 & - & - & - & - & 45 & - & 45 & 45 & 489 \\
crikey & - & 18 & 18 & 18 & - & 18 & 18 & 18 & - & 18 & 18 & 212 \\
factcheckni & - & 36 & 36 & 36 & - & - & - & 36 & - & - & 36 & 151 \\
factcheckorg & farg & 512 & 512 & 512 & 512 & 512 & 512 & 512 & 512 & 512 & 512 & 8282 \\
factly & - & 77 & 77 & - & - & - & - & 77 & - & - & 77 & 658 \\
factscan & - & 115 & 115 & - & 115 & - & - & - & 115 & 115 & 115 & 1138 \\
fullfact & - & 336 & 336 & 336 & - & 336 & - & 336 & - & 336 & 336 & 3838 \\
gossipcop & goop & 2947 & 2947 & - & - & 2947 & - & 2947 & - & 2947 & 2947 & 12583 \\
hoaxslayer & hoer & 1310 & 1310 & - & - & 1310 & - & 1310 & - & 1310 & 1310 & 14499 \\
huffingtonpostca & huca & 38 & 38 & - & 38 & 38 & - & 38 & 38 & 38 & 38 & 78 \\
leadstories & - & 1547 & 1547 & - & - & 1547 & - & 1547 & - & 1547 & 1547 & 12015 \\
mprnews & mpws & 49 & 49 & - & - & 49 & - & 49 & - & 49 & 49 & 319 \\
nytimes & - & 17 &  17 & - & - & 17 & - & 17 & - & 17 & 17 & 271 \\
observatory & obry & 60 & 60 & - & - & 60 & - & 60 & - & 60 & 60 & 592 \\
pandora & para & 225 & 225 & 225 & 225 & 225 & - & 225 & - & 225 & 225 & 114 \\
pesacheck & peck & 67 & 67 & - & - & 67 & - & 67 & - & 67 & 67 & 521 \\
politico & - & 102 & 102 & - & - & 102 & - & 102 & - & 102 & 102 & 150 \\
politifact\_promise & pose & 1361 & 1361 & 1361 & 1361 & - & - & 1361 & - & 1361 & 1361 & 6279 \\
politifact\_stmt & pomt & 15390 & 15390 & - & 15390 & - & - & - & 15390 & 15390 & 15390 & 78543 \\
politifact\_story & - & 5460 & - & - & - & 5460 & - & - & - & 5460 & 5460 & 24836 \\
radionz & ranz & 32 & 32 & 32 & 32 & - & - & 32 & 32 & 32 & 32 & 44 \\
snopes & snes & 6457 & 6457 & 6457 & - & 6457 & - & 6457 & - & 6457 & 6457 & 46735 \\
swissinfo & - & 20 & 20 & 20 & 20 & 20 & - & 20 & - & 20 & 20 & 40 \\
theconversation & - & 62 & 62 & 62 & 62 & 62 & 62 & 62 & - & 62 & 62 & 723 \\
theferret & thet & 81 & 81 & 81 & 81) & - & - & 81 & - & 81(81) & 81 & 885 \\
theguardian & - & 155 & 155 & 155 & - & 155 & - & 155 & - & 155 & 155 & 2600 \\
thejournal & thal & 179 & 179 & - & - & - & - & 179 & - & 179 & 179 & 2375 \\
truthorfiction & tron & 3674 & 3674 & 3674 & - & - & 3674 & 3674 & - & 3674 & 3674 & 8268 \\
verafiles & vees & 509 & 509 & - & - & - & 509 & 509 & - & 509 & 509 & 23 \\
voiceofsandiego & vogo & 660 & 660 & - & - & - & - & 660 & - & 660 & 660 & 2352 \\
washingtonpost & wast & 227 & 227 & - & 227 & 227 & - & 227 & - & 227 & 227 & 2470 \\
wral & - & 20 & 20 & - & - & 20 & 20 & 20 & - & 20 & 20 & 355 \\
zimfact & - & 21 & 21 & 21 & 21 & 21 & - & 21 & - & 21 & 21 & 179 \\
\midrule
Total & & 43837 & 43837 & 43837 & 43837 & 43837 & 43837 & 43837 & 43837 & 43837 & 43837 & 260330 \\
\bottomrule
\end{tabular}
\caption{\label{tb:stats4all} Summary statistics for claim collection. `Domain' indicates the domain name used for the veracity prediction experiments, `--' indicates that the website was not used due to missing or insufficient claim labels, see Section \ref{ss:evidence}.}
\end{sidewaystable*}

\end{document}